\documentclass[10pt,twocolumn,letterpaper]{article}

\usepackage{cvpr}
\usepackage{times}
\usepackage{epsfig}
\usepackage{graphicx}

\usepackage{array, multirow}
\usepackage{algorithm}
\usepackage{algorithmic}
\usepackage{mathrsfs} 
\usepackage{amsmath,bm}
\usepackage{amsthm,amssymb,amsfonts}
\usepackage{pifont}
\usepackage{color}
\usepackage{mdwlist}
\usepackage{paralist}

\usepackage{soul}
\usepackage{url}
\usepackage[small]{caption}
\usepackage{subfigure}
\usepackage{setspace}
\usepackage{textcomp}
\usepackage{bbding}
\usepackage{booktabs}
\usepackage{threeparttable}
\usepackage{authblk}

\def\comment#1{{}}
\def\eg{{\em e.g.}}
\def\ie{{\em i.e.}}
\def\etal{{\em et al.}}

\usepackage[pagebackref=true,breaklinks=true,letterpaper=true,colorlinks,bookmarks=false]{hyperref}

\cvprfinalcopy 

\def\cvprPaperID{2298} 

\ifcvprfinal\fi
\begin{document}
	
\title{Attention Convolutional Binary Neural Tree for \\ Fine-Grained Visual Categorization}

\author{Ruyi Ji$^{1,2,\ast}$, Longyin Wen$^{3,}$\thanks{Contributed equally.}, Libo Zhang$^{1,}$\thanks{Corresponding author: Libo Zhang (libo@iscas.ac.cn). This work was supported by the National Natural Science Foundation of China, Grant No. 61807033, the Key Research Program of Frontier Sciences, CAS, Grant No. ZDBS-LY-JSC038. Libo Zhang was supported by Youth Innovation Promotion Association, CAS (2020111), and Outstanding Youth Scientist Project of ISCAS.}, Dawei Du$^{4}$, \authorcr Yanjun Wu$^1$, Chen Zhao$^1$, Xianglong Liu$^5$, Feiyue Huang$^6$\\
$^1$State Key Laboratory of Computer Science, Institute of Software Chinese Academy of Sciences\\
$^2$University of Chinese Academy of Sciences\\
$^3$JD Digits \quad $^4$University at Albany, State University of New York\\
$^5$Beihang University \quad
$^6$Tencent Youtu Lab\\
{\tt\small \{ruyi2017, libo, yanjun, zhaochen\}@iscas.ac.cn, \par longyin.wen@jd.com, cvdaviddo@gmail.com}}

\maketitle

\begin{abstract}
Fine-grained visual categorization (FGVC) is an important but challenging task due to high intra-class variances and low inter-class variances caused by deformation, occlusion, illumination, etc. An attention convolutional binary neural tree architecture is presented to address those problems for weakly supervised FGVC. Specifically, we incorporate convolutional operations along edges of the tree structure, and use the routing functions in each node to determine the root-to-leaf computational paths within the tree. The final decision is computed as the summation of the predictions from leaf nodes. The deep convolutional operations learn to capture the representations of objects, and the tree structure characterizes the coarse-to-fine hierarchical feature learning process. In addition, we use the attention transformer module to enforce the network to capture discriminative features. Several experiments on the CUB-200-2011, Stanford Cars and Aircraft datasets demonstrate that our method performs favorably against the state-of-the-arts. Code can be found
at \url{https://isrc.iscas.ac.cn/gitlab/research/acnet}.
\end{abstract}

\section{Introduction}
Fine-Grained Visual Categorization (FGVC) aims to distinguish subordinate objects categories, such as different species of birds \cite{report-wahcub_200_2011,DBLP:conf/iccv/ZhengFML17}, and flowers \cite{DBLP:conf/wacv/AngelovaZL13}. The high intra-class and low inter-class visual variances caused by deformation, occlusion, and illumination, make FGVC to be a highly challenging task.

\begin{figure}[t]
\centering
\includegraphics[width=\linewidth]{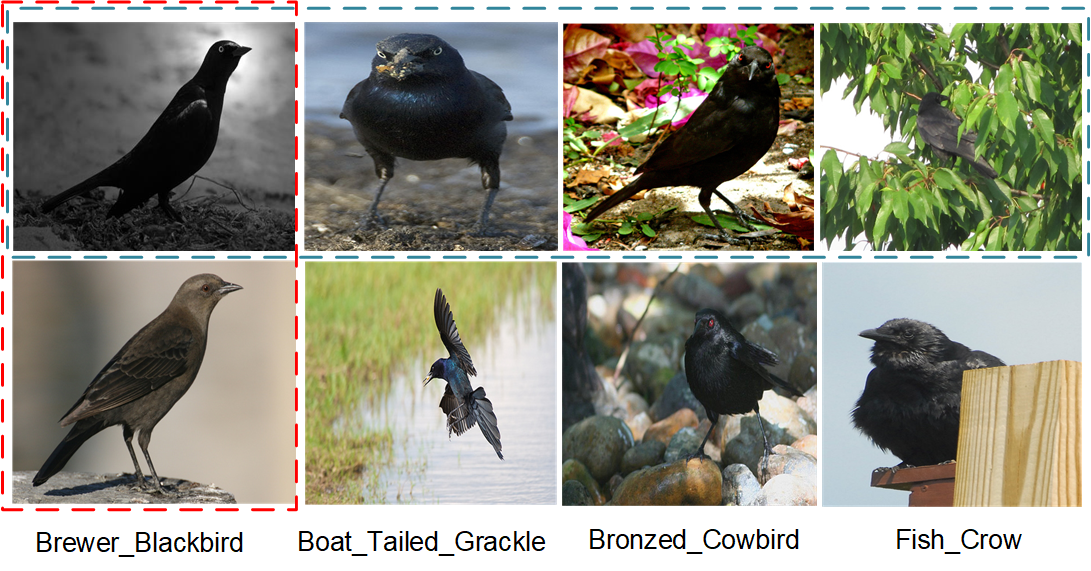}
\vspace{-4mm}
\caption{Exemplars of fine-gained visual categorization. FGVC is challenging due to two reasons: (a) high intra-class variances:  the birds belonging to the same category usually present significant different appearance, such as illumination variations (the 1st column), view-point changes (the 2nd column), clutter background (the 3rd column) and occlusion (the 4th column); (b) low inter-class variances: the birds in different columns belong to different categories, but share similar appearance in the same rows.}
\label{fig:intro}
\end{figure}

Recently, the FGVC task is dominated by the convolutional neural network (CNN) due to its amazing classification performance. Some methods \cite{DBLP:conf/iccv/LinRM15,DBLP:journals/corr/abs-1810-06058} focus on extracting discriminative subtle parts for accurate results. However, it is difficult for a single CNN model to describe the differences between subordinate classes (see Figure \ref{fig:intro}). In \cite{DBLP:journals/corr/PengHZ17}, the object-part attention model is proposed for FGVC, which uses both object and part attentions to exploit the subtle and local differences to distinguish subcategories. It demonstrates the effectiveness of using multiple deep models concentrating on different object regions in FGVC.

Inspired by \cite{DBLP:conf/icml/TannoAACN19}, we design an attention convolutional binary neural tree architecture (ACNet) for weakly supervised FGVC. It incorporates convolutional operations along the edges of the tree structure, and use the routing functions in each node to determine the root-to-leaf computational paths within the tree as deep neural networks. This designed architecture makes our method inherits the representation learning ability of the deep convolutional model, and the coarse-to-fine hierarchical feature learning process. In this way, different branches in the tree structure focus on different local object regions for classification. The final decision is computed as the summation of the predictions from all leaf nodes. Meanwhile, we use the attention transformer to enforce the tree network to capture discriminative features for accurate results. The negative log-likelihood loss is adopted to train the entire network in an end-to-end fashion by stochastic gradient descent with back-propagation.

Notably, in contrast to the work in \cite{DBLP:conf/icml/TannoAACN19} adaptively growing the tree structure in learning process, our method uses a complete binary tree structure with the pre-specified depth and the soft decision scheme to learn discriminative features in each root-to-leaf path, which avoids the pruning error and reduces the training time. In addition, the attention transformer module is used to further help our network to achieve better performance. Several experiments are conducted on the CUB-200-2011 \cite{report-wahcub_200_2011}, Stanford Cars \cite{DBLP:conf/iccvw/Krause0DF13}, and Aircraft \cite{maji13fine-grained} datasets, demonstrating the favorable performance of the proposed method compared to the state-of-the-art methods. We also carried out the ablation study to comprehensively understand the influences of different components in the proposed method.

The main contributions of this paper are summarized as follows. (1) We propose a new attention convolutional binary neural tree architecture for FGVC. (2) We introduce the attention transformer to facilitate coarse-to-fine hierarchical feature learning in the tree network. (3) Extensive experiments on three challenging datasets (\ie, CUB-200-2011, Stanford Cars, and Aircraft) show the effectiveness of our method.

\section{Related Works}
{\flushleft {\bf Deep supervised methods.}} Some algorithms \cite{DBLP:conf/eccv/ZhangDGD14,DBLP:journals/corr/LiuXWL16,DBLP:conf/cvpr/HuangXTZ16,DBLP:conf/cvpr/ZhangXEHZEM16} use object annotations or even dense part/keypoint annotations to guide the training of deep CNN model for FGVC. Zhang \etal \cite{DBLP:conf/eccv/ZhangDGD14} propose to learn two detectors, \ie, the whole object detector and the part detector, to predict the fine-grained categories based on the pose-normalized representation. Liu \etal \cite{DBLP:journals/corr/LiuXWL16} propose a fully convolutional attention networks that glimpses local discriminative regions to adapte to different fine-grained domains. The method in \cite{DBLP:conf/cvpr/HuangXTZ16} construct the part-stacked CNN architecture, which explicitly explains the fine-grained recognition process by modeling subtle differences from object parts. In \cite{DBLP:conf/cvpr/ZhangXEHZEM16}, the proposed network consists of detection and classification sub-networks. The detection sub-network is used to generate small semantic part candidates for detection; while the classification sub-network can extract features from parts detected by the detection sub-network. However, these methods rely on labor-intensive part annotations, which limits their applications in real scenarios.

{\flushleft {\bf Deep weakly supervised method.}} To that end, more recent methods \cite{DBLP:conf/iccv/ZhengFML17,DBLP:conf/cvpr/FuZM17,DBLP:conf/eccv/SunYZD18,DBLP:conf/cvpr/WangMD18} only require image-level annotations. Zheng~\etal~\cite{DBLP:conf/iccv/ZhengFML17} introduce a multi-attention CNN model, where part generation and feature learning process reinforce each other for accurate results. Fu~\etal~\cite{DBLP:conf/cvpr/FuZM17} develop a recurrent attention module to recursively learn discriminative region attention and region-based feature representation at multiple scales in a mutually reinforced way. Recently, Sun~\etal~\cite{DBLP:conf/eccv/SunYZD18} regulate multiple object parts among different input images by using multiple attention region features of each input image. In \cite{DBLP:conf/cvpr/WangMD18}, a bank of convolutional filters is learned to capture class-specific discriminative patches, through a novel asymmetric multi-stream architecture with convolutional filter supervision. However, the aforementioned methods merely integrate the attention mechanism in a single network, affecting their performance.

{\flushleft {\bf Decision tree.}} Decision tree is an effective algorithm for classification task. It selects the appropriate directions based on the characteristic of feature. The inherent ability of interpretability makes it as promising direction to pose insight into internal mechanism in deep learning. Xiao \cite{DBLP:journals/corr/abs-1712-05934} propose the principle of fully functioned neural graph and design neural decision tree model for categorization task. Frosst and Hinton \cite{DBLP:journals/corr/abs-1711-09784} develop a deep neural decision tree model to understand decision mechanism for particular test case in a learned network. Tanno~\etal~\cite{DBLP:conf/icml/TannoAACN19} propose the adaptive neural trees that incorporates representation learning into edges, routing functions and leaf nodes of a decision tree. In our work, we integrate the decision tree with neural network to implement sub-branch selection and representation learning simultaneously.

{\flushleft {\bf Attention mechanism.}}
Attention mechanism has played an important role in deep learning to mimic human visual mechanism. In \cite{DBLP:journals/corr/ZagoruykoK16a}, the attention is used to make sure the student model focuses on the discriminative regions as teacher model does. In \cite{DBLP:journals/corr/abs-1804-02391}, the cascade attention mechanism is proposed to guide the different layers of CNN and concatenate them to gain discriminative representation as the input of final linear classifier. Hu~\etal~\cite{DBLP:conf/cvpr/HuSS18} apply the attention mechanism from aspect of channels and allocate the different weights according to the contribution of each channel. The CBAM module in \cite{DBLP:journals/corr/abs-1807-06521} combines space region attentions with feature map attentions. In contrast to the aforementioned methods, we apply the attention mechanism on each branch of the tree architecture to sake the discriminative regions for classification. 

\begin{figure*}[t]
\centering
\includegraphics[width=\linewidth]{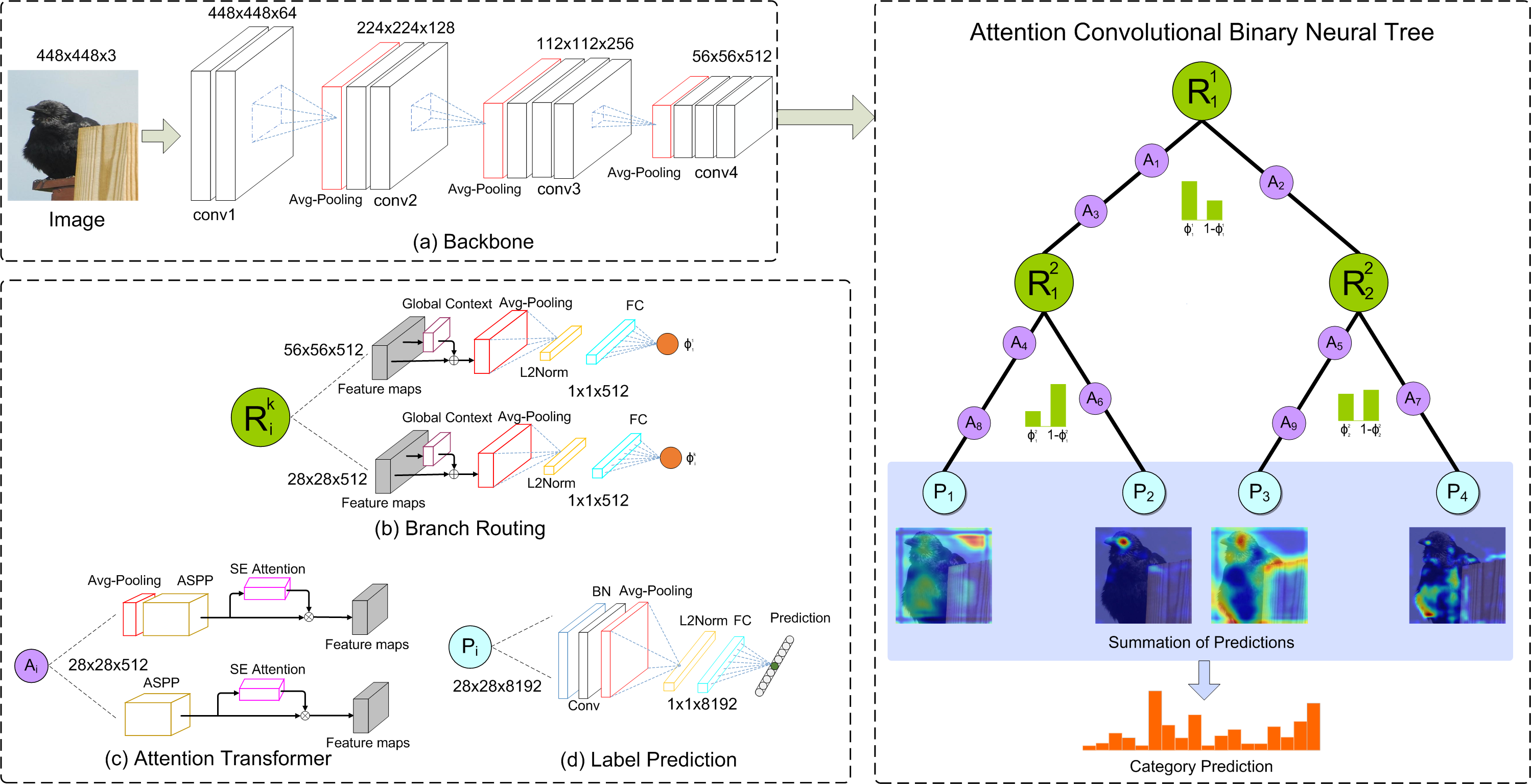}
\vspace{-4mm}
\caption{The overview of our ACNet model, formed by (a) the backbone network module, (b) the branch routing module, (c) the attention transformer module, and (d) the label prediction module. We show an example image in \textit{Fish\_Crow}. Best visualization in color.}
\label{fig:structure}
\end{figure*}

\section{Attention Convolutional Binary Neural Tree}
Our ACNet model aims to classify each object sample in ${\it X}$ to sub-categories, \ie, assign each sample in ${\it X}$ with the category label ${\it Y}$, which consists of four modules, \ie, the backbone network, the branch routing, the attention transformer, and the label prediction modules, shown in Figure \ref{fig:structure}. We define the ACNet as a pair $(\mathbb{T}, \mathbb{O})$, where $\mathbb{T}$ defines the topology of the tree, and $\mathbb{O}$ denotes the set of operations along the edges of $\mathbb{T}$. Notably, we use the full binary tree $\mathbb{T}=\{{\cal V}, {\cal E}\}$, where ${\cal V}=\{v_1, \cdots, v_{n}\}$ is the set of nodes, $n$ is the total number of nodes, and ${\cal E}=\{e_1, \cdots, e_{k}\}$ is the set of edges between nodes, $k$ is the total number of edges. Since we use the full binary tree $\mathbb{T}$, we have $n=2^{h}-1$ and $k=2^{h}-2$, where $h$ is the height of $\mathbb{T}$. Each node in $\mathbb{T}$ is formed by a routing module determining the sending path of samples, and the attention transformers are used as the operations along the edges.

Meanwhile, we use the asymmetrical architecture in the fully binary tree $\mathbb{T}$, \ie, two attention transformers are used in the left edge, and one attention transformer is used in the right edge. In this way, the network is able to capture the different scales of features for accurate results. The detail architecture of our ACNet model is described as follows.

\subsection{Architecture}
{\noindent {\bf Backbone network module}}. Since the discriminative regions in fine-grained categories are highly localized \cite{DBLP:conf/cvpr/WangMD18}, we need to use a relatively small receptive field of the extracted features by constraining the size and stride of the convolutional filters and pooling kernels. The truncated network is used as the backbone network module to extract features, which is pre-trained on the ILSVRC CLS-LOC dataset \cite{DBLP:journals/ijcv/RussakovskyDSKS15}. Similar to \cite{DBLP:conf/eccv/SunYZD18}, we use the input image size $448\times448$ instead of the default $224\times224$. Notably, ACNet can also work on other pre-trained networks, such as ResNet \cite{DBLP:conf/cvpr/HeZRS16} and Inception V2 \cite{DBLP:conf/icml/IoffeS15}. In practice, we use VGG-16 \cite{DBLP:journals/corr/SimonyanZ14a} (retaining the layers from conv1\_1 to conv4\_3) and ResNet-50 \cite{DBLP:conf/cvpr/HeZRS16} (retaining the layers from res\_1 to res\_4) networks as the backbone in this work. 

{\noindent {\bf Branch routing module}}. As described above, we use the branch routing module to determine which child (\ie, left or right child) the samples would be sent to. Specifically, as shown in Figure \ref{fig:structure}(b), the $i$-th routing module ${\cal R}_{i}^{k}(\cdot)$ at the $k$-th layer uses one convolutional layer with the kernel size $1\times1$, followed by a global context block \cite{DBLP:journals/corr/abs-1904-11492}. The global context block is an improvement of the simplified non-local (NL) block \cite{DBLP:conf/cvpr/0004GGH18} and Squeeze-Excitation (SE) block \cite{DBLP:conf/cvpr/HuSS18}, which shares the same implementation with the simplified NL block on the context modeling and fusion steps, and shares the transform step with the SE block. In this way, the context information is integrated to better describe the objects. After that, we use the global average pooling \cite{DBLP:journals/corr/LinCY13}, element-wise square-root and L2 normalization \cite{DBLP:conf/bmvc/LinM17}, and a fully connected layer with the sigmoid activation function to produce a scalar value in $[0,1]$ indicating the probability of samples being sent to the left or right sub-branches. Let $\phi_{i}^{k}(x_j)$ denote the output probability of the $j$-th sample $x_j\in{\it X}$ being sent to the right sub-branch produced by the branch routing module ${\cal R}_{i}^{k}(x_j)$, where $\phi_i^{k}(x_j)\in[0, 1]$, $i=1,\cdots,2^{k-1}$. Thus, we have the probability of the sample $x_j\in{\it X}$ being sent to the left sub-branch to be $1-\phi_i^{k}(x_j)$. If the probability $\phi_{i}^{k}(x_j)$ is larger than $0.5$, we prefer the left path instead of the right one; otherwise, the left branch dominates the final decision. 

\begin{figure}[t]
\centering
\includegraphics[width=\linewidth]{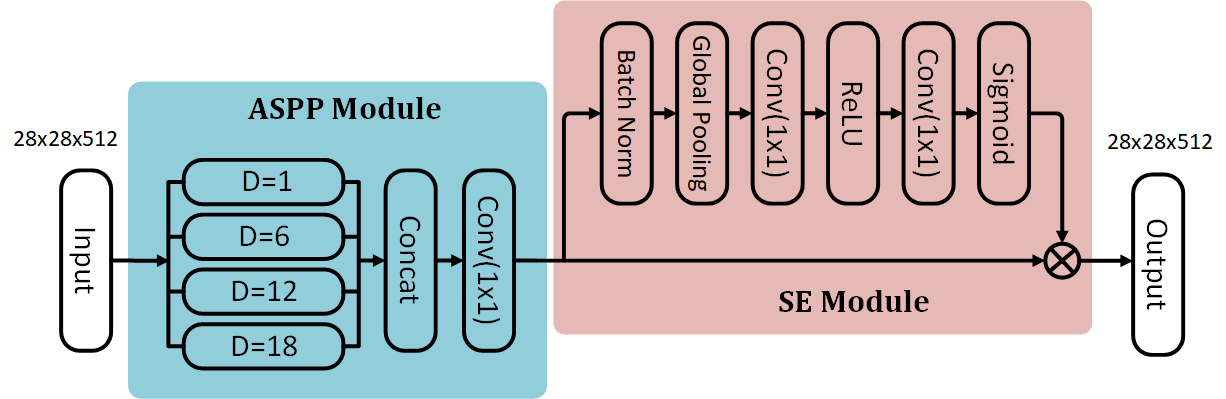}
\caption{The architecture of the attention transformer module.}
\vspace{-2mm}
\label{fig:attn-transformer}
\end{figure}

{\noindent {\bf Attention transformer}.} The attention transformer module is used to enforce the network to capture discriminative features, see Figure \ref{fig:attn-transformer}. According to the fact that the empirical receptive field is much smaller than theoretical receptive field in deep networks \cite{DBLP:journals/corr/LiuRB15}, the discriminative representation should be formed by larger receptive field in new-added layers of our proposed tree structure. To this end, we intergate the Atrous Spatial Pyramid Pooling (ASPP) module \cite{DBLP:journals/pami/ChenPKMY18} into the attention transformer. Specifically, ASPP module provides different feature maps with each characterized by a different scale/receptive field and an attention module. Then, multi-scale feature maps are generated by four parallel dilated convolutions with different dilated rates, \ie, $1, 6, 12, 18$. Following the parallel dilated convolution layers, the concatenated feature maps are fused by one convolutional layer with kernel $1\times1$ and stride $1$. Following the ASPP module, we insert an attention module, which generates a channel attention map with the size $\mathbb{R}^{{\it C}\times1\times1}$ using a batch normalization (BN) layer \cite{DBLP:conf/icml/IoffeS15}, a global average pooling (GAP) layer, a fully connected (FC) layer and ReLU activation function, and a FC layer and sigmoid function. In this way, the network is guided to focus on meaningful features for accurate results.

{\noindent {\bf Label prediction}.} For each leaf node in our ACNet model, we use the label prediction module ${\cal P}_i$ (\ie, $i=1,\cdots, 2^{h-1}$) to predict the subordinate category of the object ${\it x}_j$, see Figure \ref{fig:structure}. Let ${\it r}_i^{k}(x_j)$ to be the accumulated probability of the object $x_j$ passing from the root node to the $i$-th node at the $k$-th layer. For example, if the root to the node ${\cal R}_{i}^{k}(\cdot)$ path on the tree is ${\cal R}_{1}^{1}, {\cal R}_{1}^{2}, \cdots, {\cal R}_{1}^{k}$, \ie, the object $x_j$ is always sent to the left child, we have ${\it r}_{i}^{k}(x_j) = \prod_{i=1}^{k}\phi_{1}^{i}(x_j)$. As shown in Figure \ref{fig:structure}, the label prediction module is formed by a batch normalization layer, a convolutional layer with kernel size $1\times1$, a max-pooling layer, a sqrt and L2 normalization layer, and a fully connected layer. Then, the final prediction  ${\cal C}({\it x}_j)$ of the $j$-th object ${\it x}_j$ is computed as the summation of all leaf predictions multiplied with the accumulated probability generated by the passing branch routing modules, \ie, ${\cal C}(x_j)=\sum_{i=1}^{2^{h-1}}{\cal P}_i(x_j){\it r}_i^{h}(x_j)$. We would like to emphasize that $\|{\cal C}(x_j)\|_1=1$, \ie, the summation of confidences of $x_j$ belonging to all subordinate classes equal to $1$,
\begin{equation}
\begin{array}{cl}
\|{\cal C}({\it x}_j)\|_1 = \|\sum_{i=1}^{2^{h-1}}{{\cal P}_i({\it x}_j)}{\it r}_i^{h}({\it x}_j)\|_1 = 1,
\end{array}
\end{equation}
where ${\it r}_i^{h}({\it x}_j)$ is the accumulated probability of the $i$-th node at the leaf layer. We present a short description to prove that $\|{\cal C}({\it x}_j)\|_1=1$ as follows.
\begin{proof}
Let ${\it r}_i^{k}(\cdot)$ be the accumulated probability of the $i$-th branch routing module ${\cal R}_{i}^{k}(\cdot)$ at the $k$-th layer. Thus, the accumulated probabilities of the left and right children corresponding to ${\cal R}_{i}^{k}(\cdot)$ are $r_{2i-1}^{k+1}(\cdot)$ and $r_{2i}^{k+1}(\cdot)$, respectively. At first, we demonstrate that the summation of the accumulated probabilities $r_{2i-1}^{k+1}(\cdot)$ and $r_{2i}^{k+1}(\cdot)$ is equal to the accumulated probability of their parent ${\it r}_i^{k}({\it x}_j)$. That is, 
\begin{equation}
\begin{aligned}
        &r_{2i-1}^{k+1}(x_j) +r_{2i}^{k+1}(x_j) \\
        &= \phi_{2i-1}^{k+1}(x_j) \cdot r_i^{k}(x_j) + \phi_{2i}^{k+1}(x_j) \cdot r_i^{k}(x_j)\\
        &= \phi_{2i-1}^{k+1}(x_j) \cdot r_i^{k}(x_j) + \big(1-\phi_{2i-1}^{k+1}(x_j)\big) \cdot r_i^{k}(x_j)\\
        &= r_i^{k}(x_j).
\end{aligned}
\end{equation}
Meanwhile, due to the fully binary tree $\mathbb{T}$ in ACNet, we have $\sum_{i=1}^{2^{h-1}} r_i^{h}(x_j) = \sum_{i=1}^{2^{h-2}}\big( r_{2i-1}^{h}(x_j) +r_{2i}^{h}(x_j) \big)$. We can further get $\sum_{i=1}^{2^{h-1}} r_i^{h}(x_j) =  \sum_{i=1}^{2^{h-2}} r_i^{h-1}(x_j)$. This process is carried out iteratively, and we have $\sum_{i=1}^{2^{h-1}} r_i^{h}(x_j) = \dots = r_1^{1}(x_j) = 1$. In addition, since the category prediction $\mathcal{P}_i(x_i)$ is generated by the softmax layer (see Figure \ref{fig:structure}), we have $\|\mathcal{P}_i(x_j)\|_1 = 1$. Thus,  
\begin{equation}
\begin{array}{cl}
        \|\mathcal{C}(x_j)\|_1 &= \|\sum_{i=1}^{2^{h-1}} \mathcal{P}_i(x_j)r_i^{h}(x_j)\|_1 \\
                                          &= \sum_{i=1}^{2^{h-1}} \|\mathcal{P}_i(x_j)\|_1r_i^{h}(x_j) = 1.
\end{array}
\end{equation}
\end{proof}
As shown in Figure \ref{fig:structure}, when occlusion happens, ACNet can still localize some discriminative object parts and context information of the bird. Although high intra-class visual variances always happen in FGVC, ACNet uses a coarse-to-fine hierarchical feature learning process to exploit the discriminative feature for classification. In this way, different branches in the tree structure focus on different fine-grained object regions for accurate results.  

\subsection{Training}
{\flushleft \textbf{Data augmentation}.} In the training phase, we use the cropping and flipping operations to augment data to construct a robust model to adapt to variations of objects. That is, we first rescale the original images such that  their shorter side is $512$ pixels. After that, we randomly crop the patches with the size $448\times448$, and randomly flip them to generate the training samples.

{\flushleft \textbf{Loss function}.} The loss function for our ACNet is formed by two parts, \ie, the loss for the predictions of leaf nodes, and the loss for the final prediction, computed by the summation over all predictions from the leaf nodes. That is,
\begin{equation}
{\cal L} = {\it L}\big({\cal C}({\it x}_j), {\it y}^{\ast}\big) + \sum_{i=1}^{2^{h-1}}{\it L}\big({\cal P}_{i}({\it x}_j), {\it y}^{\ast}\big),
\end{equation}
where $h$ is the height of the tree $\mathbb{T}$, ${\it L}\big({\cal C}({\it x}_j), {\it y}^{\ast}\big)$ is the negative logarithmic likelihood loss of the final prediction ${\cal C}({\it x}_j)$ and the ground truth label ${\it y}^{\ast}$, and ${\it L}\big({\cal P}_{i}({\it x}_j), {\it y}^{\ast}\big)$ is the negative logarithmic likelihood loss of the $i$-th leaf prediction and the ground truth label ${\it y}^{\ast}$.

{\flushleft \textbf{Optimization}.}
The backbone network in our ACNet method is pre-trained on the ImageNet dataset. Besides, the ``xavier'' method \cite{DBLP:journals/jmlr/GlorotB10} is used to randomly initialize the parameters of the additional convolutional layers. The entire training process is formed by two stages. 
\begin{itemize}
\item For the first stage, the parameters in the truncated VGG-16 network are fixed, and other parameters are trained with $60$ epochs. The batch size is set to $24$ in training with the initial learning rate $1.0$. The learning rate is gradually divided by $4$ at the $10$-th, $20$-th, $30$-th, and $40$-th epochs. 
\item In the second stage, we fine-tune the entire network for $200$ epochs. We use the batch size $16$ in training with the initial learning rate $0.001$. The learning rate is gradually divided by $10$ at the $30$-th, $40$-th, and $50$-th epochs. 
\end{itemize}
We use the SGD algorithm to train the network with $0.9$ momentum, and $0.000005$ weight decay in the first stage and $0.0005$ weight decay in the second stage.

\begin{table}[t]
    \centering
    \small { \setlength{\tabcolsep}{1.0pt}
    \caption{The fine-grained classification results on the CUB-200-2011 dataset \cite{report-wahcub_200_2011}.}
    \label{tab:birds}    
    \begin{tabular}{ccccpppp}
    	\toprule
    	                   Method                     &   Backbone   & Annotation & Top-1 Acc. (\%) \\ \midrule
    	   FCAN \cite{DBLP:journals/corr/LiuXWL16}    &  ResNet-50   & \checkmark &      84.7       \\
    	     B-CNN \cite{DBLP:conf/iccv/LinRM15}      &    VGG-16    & \checkmark &      85.1       \\
    	SPDA-CNN \cite{DBLP:conf/cvpr/ZhangXEHZEM16}  &   CaffeNet   & \checkmark &      85.1       \\
    	PN-CNN \cite{DBLP:journals/corr/BransonHBP14} &   Alex-Net   & \checkmark &      85.4       \\ \midrule
    	  STN \cite{DBLP:conf/nips/JaderbergSZK15}    &  Inception   &  $\times$  &      84.1       \\
    	     B-CNN \cite{DBLP:conf/iccv/LinRM15}      &    VGG-16    &  $\times$  &      84.0       \\
    	   CBP \cite{DBLP:journals/corr/GaoBZD15}     &    VGG-16    &  $\times$  &      84.0       \\
    	   LRBP \cite{DBLP:journals/corr/KongF16}     &    VGG-16    &  $\times$  &      84.2       \\
    	   FCAN \cite{DBLP:journals/corr/LiuXWL16}    &  ResNet-50   &  $\times$  &      84.3       \\
    	     RA-CNN \cite{DBLP:conf/cvpr/FuZM17}      &    VGG-19    &  $\times$  &      85.3       \\
    	     HIHCA \cite{DBLP:conf/iccv/CaiZZ17}      &    VGG-16    &  $\times$  &      85.3       \\
    	 Improved B-CNN \cite{DBLP:conf/bmvc/LinM17}  &    VGG-16    &  $\times$  &      85.8       \\
    	BoostCNN \cite{DBLP:conf/bmvc/MoghimiBSYVL16} &    VGG-16    &  $\times$  &      86.2       \\
    	     KP \cite{DBLP:conf/cvpr/CuiZWLLB17}      &    VGG-16    &  $\times$  &      86.2       \\
    	   MA-CNN \cite{DBLP:conf/iccv/ZhengFML17}    &    VGG-19    &  $\times$  &      86.5       \\
    	     MAMC \cite{DBLP:conf/eccv/SunYZD18}      &  ResNet-101  &  $\times$  &      86.5       \\
    	   MaxEnt \cite{DBLP:conf/nips/DubeyGRN18}    & DenseNet-161 &  $\times$  &      86.5       \\
    	            HBPASM \cite{8805063}             &  Resnet-34   &  $\times$  &      86.8       \\
    	          DCL \cite{Chen_2019_CVPR}           &    VGG-16    &  $\times$  &      86.9       \\
    	KERL w/ HR \cite{DBLP:conf/ijcai/ChenLCWL18}  &    VGG-16    &  $\times$  &      87.0       \\
    	TASN \cite{DBLP:journals/corr/abs-1903-06150} &    VGG-19    &  $\times$  &      87.1       \\
    	   DFL-CNN \cite{DBLP:conf/cvpr/WangMD18}     &  ResNet-50   &  $\times$  &      87.4       \\
    	          DCL \cite{Chen_2019_CVPR}           &  ResNet-50   &  $\times$  &      87.8       \\
    	TASN \cite{DBLP:journals/corr/abs-1903-06150} &  ResNet-50   &  $\times$  &      87.9       \\
    	                    Ours                      &    VGG-16    &  $\times$  &  \textbf{87.8}  \\
    	                    Ours                      &  ResNet-50   &  $\times$  &  \textbf{88.1}  \\ \bottomrule
    \end{tabular}}
    \vspace{-2mm}
\end{table}

\section{Experiments}
Several experiments on three FGVC datasets, \ie, CUB-200-2011 \cite{report-wahcub_200_2011}, Stanford Cars \cite{DBLP:conf/iccvw/Krause0DF13}, and Aircraft \cite{maji13fine-grained}, are conducted to demonstrate the effectiveness of the proposed method. Our method is implemented in the Caffe library \cite{DBLP:journals/corr/JiaSDKLGGD14}. All models are trained on a workstation with a 3.26 GHz Intel processor, 32 GB memory, and one Nvidia V100 GPU.

\begin{table}[t]
    \centering
    \small{ \setlength{\tabcolsep}{1.0pt}
    \caption{The fine-grained classification results on the Stanford Cars dataset \cite{DBLP:conf/iccvw/Krause0DF13}.}
    \vspace{-2mm}
    \label{tab:cars}    
    \begin{tabular}{cccc}
    	\toprule
    	                    Method                      &   Backbone   & Annotation & Top-1 Acc. (\%) \\ \midrule
    	    FCAN \cite{DBLP:journals/corr/LiuXWL16}     &  ResNet-50   & \checkmark &      91.3       \\
    	   PA-CNN \cite{DBLP:conf/cvpr/KrauseJYL15}     &    VGG-19    & \checkmark &      92.6       \\ \midrule
    	    FCAN \cite{DBLP:journals/corr/LiuXWL16}     &  ResNet-50   &  $\times$  &      89.1       \\
    	      B-CNN \cite{DBLP:conf/iccv/LinRM15}       &    VGG-16    &  $\times$  &      90.6       \\
    	    LRBP \cite{DBLP:journals/corr/KongF16}      &    VGG-16    &  $\times$  &      90.9       \\
    	      HIHCA \cite{DBLP:conf/iccv/CaiZZ17}       &    VGG-16    &  $\times$  &      91.7       \\
    	  Improved B-CNN \cite{DBLP:conf/bmvc/LinM17}   &    VGG-16    &  $\times$  &      92.0       \\
    	 BoostCNN \cite{DBLP:conf/bmvc/MoghimiBSYVL16}  &    VGG-16    &  $\times$  &      92.1       \\
    	      KP \cite{DBLP:conf/cvpr/CuiZWLLB17}       &    VGG-16    &  $\times$  &      92.4       \\
    	      RA-CNN \cite{DBLP:conf/cvpr/FuZM17}       &    VGG-19    &  $\times$  &      92.5       \\
    	    MA-CNN \cite{DBLP:conf/iccv/ZhengFML17}     &    VGG-19    &  $\times$  &      92.8       \\
    	      MAMC \cite{DBLP:conf/eccv/SunYZD18}       &  ResNet-101  &  $\times$  &      93.0       \\
    	    MaxEnt \cite{DBLP:conf/nips/DubeyGRN18}     & DenseNet-161 &  $\times$  &      93.0       \\
    	WS-DAN \cite{DBLP:journals/corr/abs-1901-09891} & Inception v3 &  $\times$  &      93.0       \\
    	    DFL-CNN \cite{DBLP:conf/cvpr/WangMD18}      &  ResNet-50   &  $\times$  &      93.1       \\
    	             HBPASM \cite{8805063}              &  Resnet-34   &  $\times$  &      93.8       \\
    	 TASN \cite{DBLP:journals/corr/abs-1903-06150}  &    VGG-19    &  $\times$  &      93.2       \\
    	 TASN \cite{DBLP:journals/corr/abs-1903-06150}  &  ResNet-50   &  $\times$  &      93.8       \\
    	           DCL \cite{Chen_2019_CVPR}            &    VGG-16    &  $\times$  &      94.1       \\
    	           DCL \cite{Chen_2019_CVPR}            &  ResNet-50   &  $\times$  &      94.5       \\
	           \hline
    	                     Ours                       &    VGG-16    &  $\times$  &  \textbf{94.3}  \\
    	                     Ours                       &  ResNet-50   &  $\times$  &  \textbf{94.6}  \\ \bottomrule
    \end{tabular}}
\end{table}

\subsection{Evaluation on the CUB-200-2011 Dataset}
The Caltech-UCSD birds dataset (CUB-200-2011) \cite{report-wahcub_200_2011} consists of $11,788$ annotated images in $200$ subordinate categories, including $5,994$ images for training and $5,794$ images for testing. The fine-grained classification results are shown in Table \ref{tab:birds}. As shown in Table \ref{tab:birds}, the best supervised method\footnote{Notably, the supervised method requires object or part level annotations, demanding significant human effort. Thus, most of recent methods focus on the weakly supervised methods, pushing the state-of-the-art weakly supervised methods surpassing the performance of previous supervised methods.}, \ie, PN-CNN \cite{DBLP:journals/corr/BransonHBP14} using both the object and part level annotations produces $85.4\%$ top-1 accuracy on the CUB-200-2011 dataset. Without part-level annotation, MAMC \cite{DBLP:conf/eccv/SunYZD18} produces $86.5\%$ top-1 accuracy using two attention branches to learn discriminative features in different regions. KERL w/ HR \cite{DBLP:conf/ijcai/ChenLCWL18} designs a single deep gated graph neural network to learn discriminative features, achieving better performance, \ie, $87.0\%$ top-1 accuracy. Compared to the state-of-the-art weakly supervised methods \cite{DBLP:conf/ijcai/ChenLCWL18,DBLP:conf/nips/DubeyGRN18,DBLP:conf/eccv/SunYZD18,DBLP:conf/cvpr/WangMD18,Chen_2019_CVPR,DBLP:journals/corr/abs-1903-06150}, our method achieves the best results with $87.8\%$ and $88.1\%$ top-1 accuracy with different backbones. This is attributed to the designed attention transformer module and the coarse-to-fine hierarchical feature learning process. 

\subsection{Evaluation on the Stanford Cars Dataset}
The Stanford Cars dataset \cite{DBLP:conf/iccvw/Krause0DF13} contains $16,185$ images from $196$ classes, which is formed by $8,144$ images for training and $8,041$ images for testing. The subordinate categories are determined by the \textit{Make}, \textit{Model}, and \textit{Year} of cars. As shown in Table \ref{tab:cars}, previous methods using part-level annotations (\ie, FCAN \cite{DBLP:journals/corr/LiuXWL16} and PA-CNN \cite{DBLP:conf/cvpr/KrauseJYL15}) only produces less than $93.0\%$ top-$1$ accuracy. The recent weakly supervised method WS-DAN \cite{DBLP:journals/corr/abs-1901-09891} employs the complex Inception V3 backbone \cite{DBLP:conf/cvpr/SzegedyVISW16} and designs the attention-guided data augmentation strategy to exploit discriminative object parts, achieving $93.0\%$ top-1 accuracy. Without using any fancy data augmentation strategy, our method achieves the best top-$1$ accuracy, \ie, $94.3\%$ with the VGG-16 backbone and $94.6\%$ with the ResNet-50 backbone. 

\begin{table}[t]
    \centering
    \small{ \setlength{\tabcolsep}{1.0pt}
    \caption{The fine-grained classification results on the Aircraft dataset \cite{maji13fine-grained}.}
    \vspace{-2mm}
    \label{tab:aircraft}    
    \begin{tabular}{cccc}
    	\toprule
    	                      Method                       &   Backbone   & Annotation & Top-1 Acc. (\%) \\ \midrule
    	     MG-CNN \cite{DBLP:conf/iccv/WangSSZXZ15}      &  ResNet-50   & \checkmark &      86.6       \\
    	     MDTP \cite{DBLP:journals/corr/WangCMD16}      &    VGG-16    & \checkmark &      88.4       \\ \midrule
    	       RA-CNN \cite{DBLP:conf/cvpr/FuZM17}         &    VGG-19    &  $\times$  &      88.2       \\
    	     MA-CNN \cite{DBLP:conf/iccv/ZhengFML17}       &    VGG-19    &  $\times$  &      89.9       \\
    	       B-CNN \cite{DBLP:conf/iccv/LinRM15}         &    VGG-16    &  $\times$  &      86.9       \\
    	       KP \cite{DBLP:conf/cvpr/CuiZWLLB17}         &    VGG-16    &  $\times$  &      86.9       \\
    	      LRBP \cite{DBLP:journals/corr/KongF16}       &    VGG-16    &  $\times$  &      87.3       \\
    	       HIHCA \cite{DBLP:conf/iccv/CaiZZ17}         &    VGG-16    &  $\times$  &      88.3       \\
    	   Improved B-CNN \cite{DBLP:conf/bmvc/LinM17}     &    VGG-16    &  $\times$  &      88.5       \\
    	  BoostCNN \cite{DBLP:conf/bmvc/MoghimiBSYVL16}    &    VGG-16    &  $\times$  &      88.5       \\
    	PC-DenseNet-161 \cite{DBLP:conf/eccv/DubeyGGRFN18} & DenseNet-161 &  $\times$  &      89.2       \\
    	     MaxEnt \cite{DBLP:conf/nips/DubeyGRN18}       & DenseNet-161 &  $\times$  &      89.7       \\
    	              HBPASM \cite{8805063}                &  Resnet-34   &  $\times$  &      91.3       \\
    	      DFL-CNN \cite{DBLP:conf/cvpr/WangMD18}       &  ResNet-50   &  $\times$  &      91.7       \\
    	            DCL \cite{Chen_2019_CVPR}              &    VGG-16    &  $\times$  &      91.2       \\
    	            DCL \cite{Chen_2019_CVPR}              &  ResNet-50   &  $\times$  &  \textbf{93.0}  \\
	            \hline
    	                       Ours                        &    VGG-16    &  $\times$  &  \textbf{91.5}  \\
    	                       Ours                        &  ResNet-50   &  $\times$  &      92.4       \\ \bottomrule
    \end{tabular}}
\end{table}

\subsection{Evaluation on the Aircraft Dataset}
The Aircraft dataset \cite{maji13fine-grained} is a fine-grained dataset of $100$ different aircraft variants formed by $10,000$ annotated images, which is divided into two subsets, \ie, the training set with $6,667$ images and the testing set with $3,333$ images. Specifically, the category labels are determined by the \textit{Model}, \textit{Variant}, \textit{Family} and \textit{Manufacturer} of airplanes. The evaluation results are presented in Table \ref{tab:aircraft}. 
Our ACNet method outperforms the most compared methods, especially with the same VGG-16 backbone. Besides, our model performs on par with the state-of-the-art method DCL \cite{Chen_2019_CVPR}, \ie, $91.2\%$ \textit{vs.} $91.5\%$ top-1 accuracy for the VGG-16 backbone and $93.0\%$ \textit{vs.} $92.4\%$ top-1 accuracy for the ResNet-50 backbone. The operations along different root-to-leaf paths in our tree architecture $\mathbb{T}$ focus on exploiting discriminative features on different object regions, which help each other to achieve the best performance in FGVC.

\begin{figure}[t]
\centering
\includegraphics[width=0.95\linewidth]{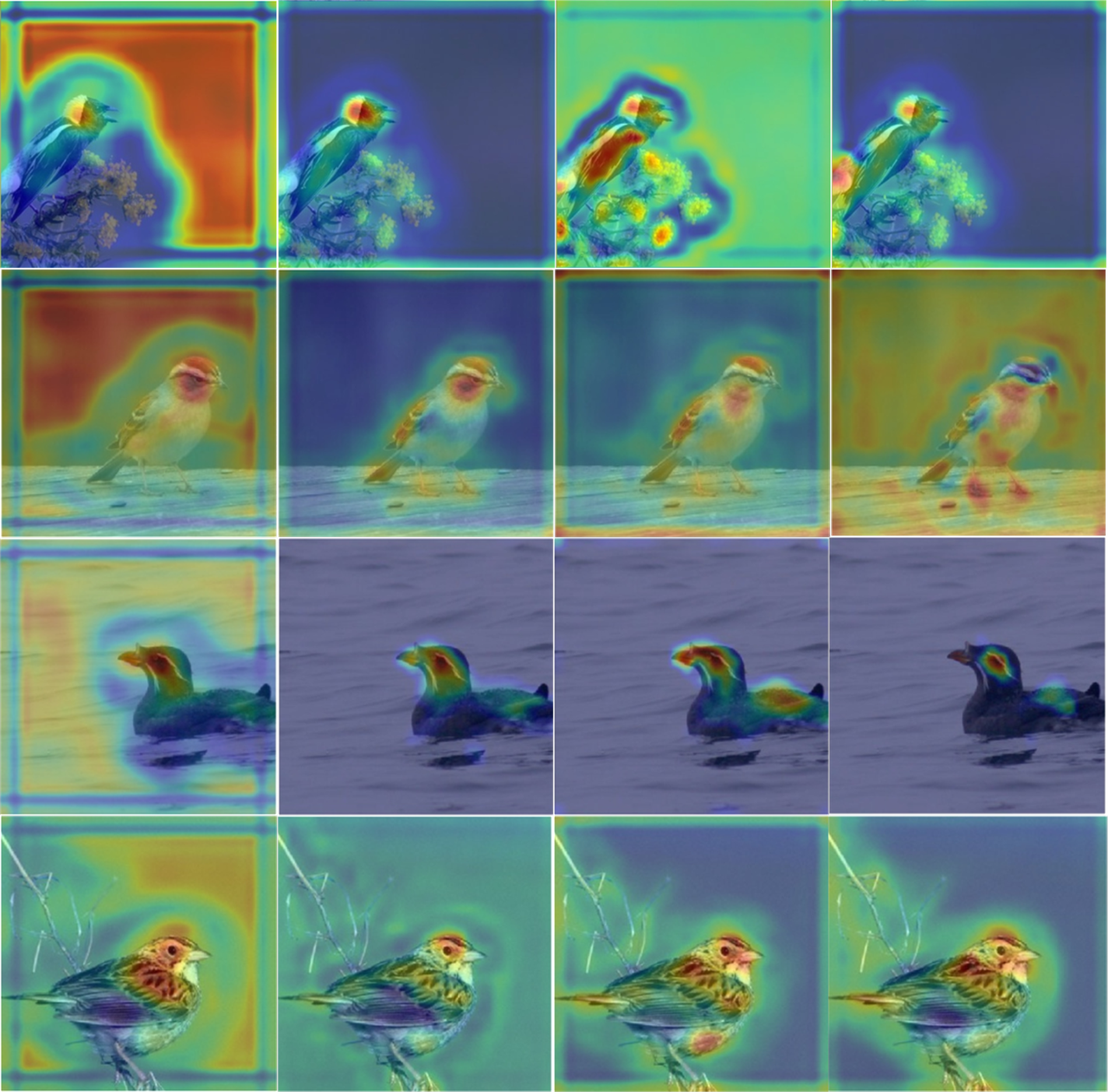}
\caption{Visualization of the responses in different leaf nodes in our ACNet method. Each column presents a response heatmap of each leaf node.}
\label{fig:attention}
\vspace{-2mm}
\end{figure}

\subsection{Ablation Study}
We study the influence of some important parameters and different components of ACNet on the CUB-200-2011 dataset \cite{report-wahcub_200_2011}. Notably, we employ the VGG-16 backbone in the experiment. The Grad-CAM method \cite{DBLP:conf/iccv/SelvarajuCDVPB17} is used to generate the heatmaps to visualize the responses of branch routing and leaf nodes.

\begin{figure}[t]
\centering
\includegraphics[width=0.95\linewidth]{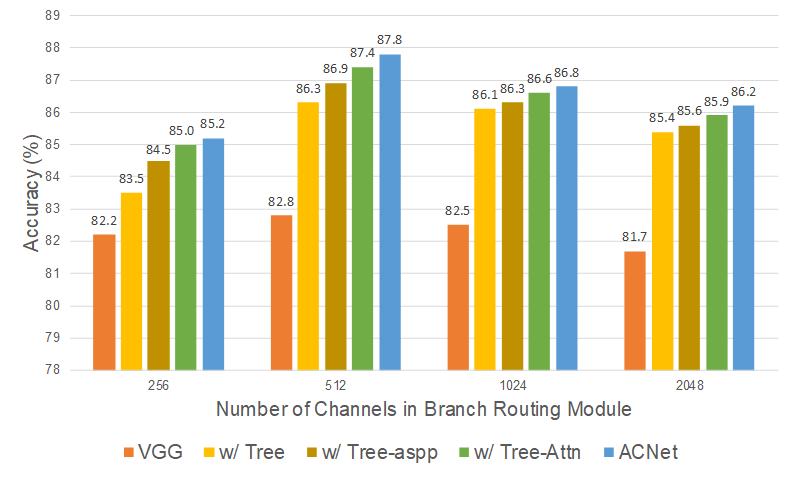}
\caption{Effect of the various components in the proposed ACNet method on the CUB-200-2011 dataset \cite{report-wahcub_200_2011}.}
\vspace{-2mm}
\label{fig:ablation}
\end{figure}

{\flushleft \textbf{Effectiveness of the tree architecture $\mathbb{T}$.}} To validate the effectiveness of the tree architecture design, we construct two variants, \ie, VGG and w/ Tree, of our ACNet method. Specifically, we construct the VGG method by only using the VGG-16 backbone network for classification, and further integrate the tree architecture to form the w/ Tree method. The evaluation results are reported in Figure \ref{fig:ablation}. We find that using the tree architecture significantly improves the accuracy, \ie, $3.025\%$ improvements in top-1 accuracy, which demonstrates the effectiveness of the designed tree architecture $\mathbb{T}$ in our ACNet method. 

\begin{figure}[h]
	\centering
	\includegraphics[width=0.95\linewidth]{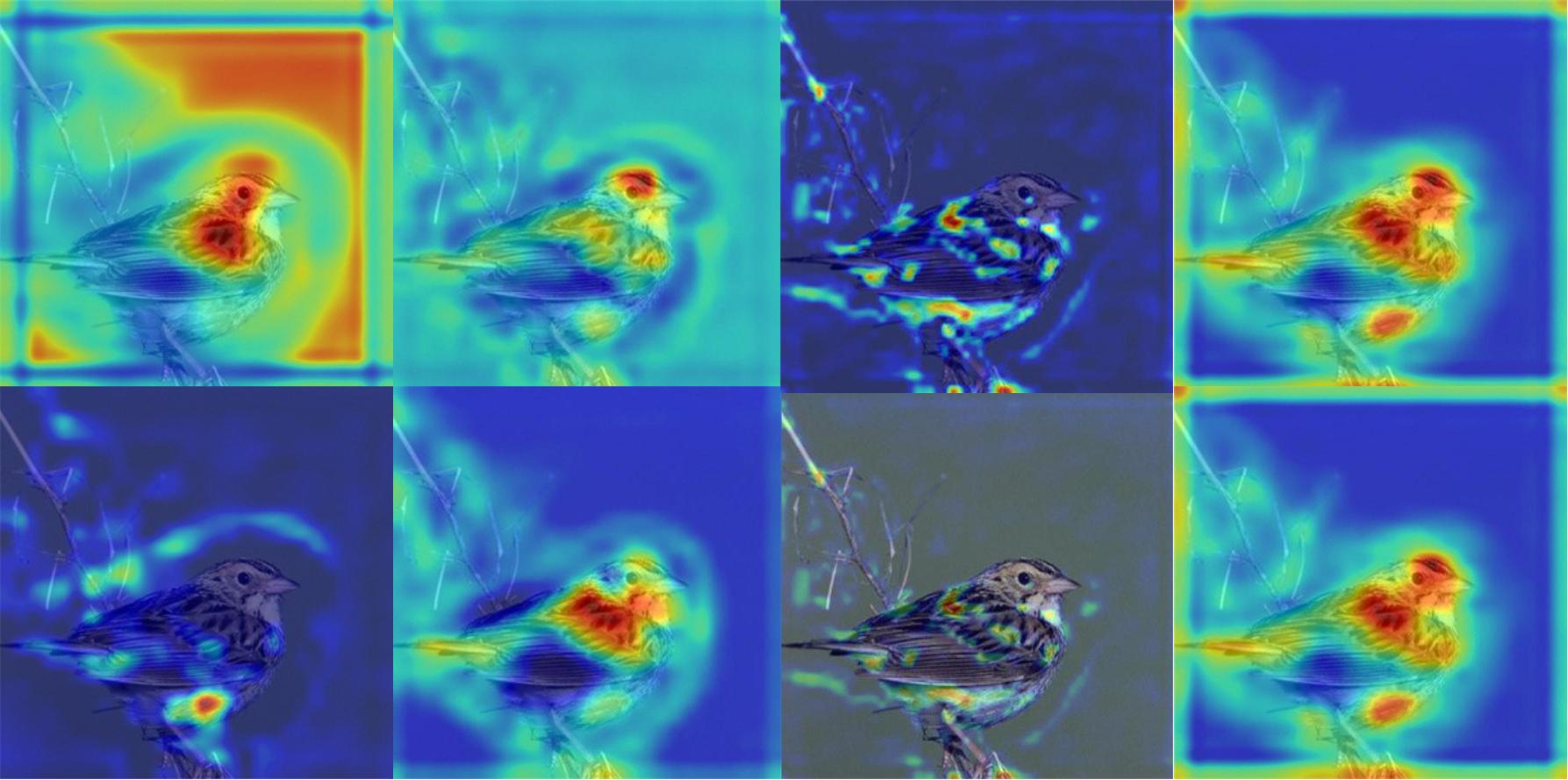}
	\caption{Responses of all leaf nodes in the tree with height of $4$.}
	\vspace{-2mm}
	\label{fig:leaf-node}
\end{figure}

{\flushleft \textbf{Height of the tree $\mathbb{T}$.}}
To explore the effect of the height of the tree $\mathbb{T}$, we construct four variants with different heights of tree in Table \ref{tab:leaf-node}. Notably, the tree $\mathbb{T}$ is degenerated to a single node when the height of the tree is set to $1$, \ie, only the backbone VGG-16 network is used in classification.  As shown in Table \ref{tab:leaf-node}, we find that our ACNet achieves the best performance (\ie, $87.8\%$ top-1 accuracy) with the height of tree equals to $3$. If we set $h\leq2$, there are limited number of parameters in our ACNet model, which are not enough to represent the significant variations of the subordinate categories. However, if we set $h=4$, too many parameters with limited number of training data cause overfitting of our ACNet model, resulting in $2.3\%$ drop in the top-$1$ accuracy. To verify our hypothesis, we visualize the responses of all leaf nodes in ACNet with the height of $4$ in Figure \ref{fig:leaf-node}. We find that some leaf nodes focus on almost the same regions (see the 3rd and 4th columns).

\begin{table*}[t]
    \begin{minipage}{0.33\linewidth}
    \centering
    { \setlength{\tabcolsep}{6.5pt}
    \caption{Effect of the height of the tree $\mathbb{T}$.}
    \vspace{-2mm}
    \label{tab:leaf-node}    
    \begin{tabular}{cc}
    	\toprule
    	Height of $\mathbb{T}$ & Top-1 Acc. (\%) \\ \midrule
    	        1          &      82.2       \\
    	        2          &      86.0       \\
    	        3          &    \bf{87.8}    \\
    	        4          &      85.5       \\ \bottomrule
    \end{tabular}}    
    \end{minipage}
   \begin{minipage}{0.33\linewidth}
    \centering
    { \setlength{\tabcolsep}{3.5pt}
    \renewcommand\arraystretch{1.6}
    \caption{Effect of tree architecture.}
     \vspace{-2mm}
    \label{tab:asymmetry}    
    \begin{tabular}{ccc}
    	\toprule
    	  Mode    & Level & Top-1 Acc. (\%) \\ \midrule
    	symmetry  &   3   &      86.2       \\
    	asymmetry &   3   &    \bf{87.8}    \\ \bottomrule
    \end{tabular}}
    \end{minipage}
    \begin{minipage}{0.33\linewidth}
    \centering
    {\setlength{\tabcolsep}{8.5pt}
    \renewcommand\arraystretch{1.55}
     \vspace{-2mm}
    \caption{Comparison between GMP and GAP.}
    \label{tab:pooling}    
    \begin{tabular}{cc}
    	\toprule
    	Pooling & Top-1 Acc. (\%) \\ \midrule
    	  GMP   &      87.2       \\
    	  GAP   &    \bf{87.8}    \\ \bottomrule
    \end{tabular}}    
    \end{minipage}
    \vspace{-2mm}
\end{table*}

{\flushleft \textbf{Effectiveness of leaf nodes.}} To analyze the effectiveness of the individual leaf node, we calculate the accuracy of individual leaf predictions with height of $3$, respectively. The accuracies of four individual leaf nodes are $85.8\%$, $86.2\%$, $86.7\%$, and $87.0\%$ on CUB-200-2011 respectively. It shows that all leaf nodes are informative and fusion of them can produce more accurate result (\ie, $87.8\%$). As shown in Figure \ref{fig:attention}, we observe that different leaf nodes concentrate on different regions of images. For example, the leaf node corresponding to the first column focuses more on the background region, the leaf node corresponding to the second column focuses more on the head region, and the other two leaf nodes are more interested in the patches of wings and tail. The different leaf nodes help each other to construct more effective model for accurate results. 

{\flushleft \textbf{Asymmetrical architecture of the tree $\mathbb{T}$.}}
To explore the architecture design in $\mathbb{T}$, we construct two variants, \ie, one uses the symmetry architecture, and another one uses the asymmetrical architecture, and set the height of the tree $\mathbb{T}$ to be $3$. The evaluation results are reported in Table \ref{tab:asymmetry}. It can be seen that the proposed method produces $86.2\%$ top-1 accuracy using the symmetrical architecture. If we use the asymmetrical architecture, the top-1 accuracy is improved $1.6\%$ to $87.8\%$. We speculate that the asymmetrical architecture is able to fuse various features with different receptive fields for better performance. 

\begin{figure}[t]
\centering
\includegraphics[width=0.95\linewidth]{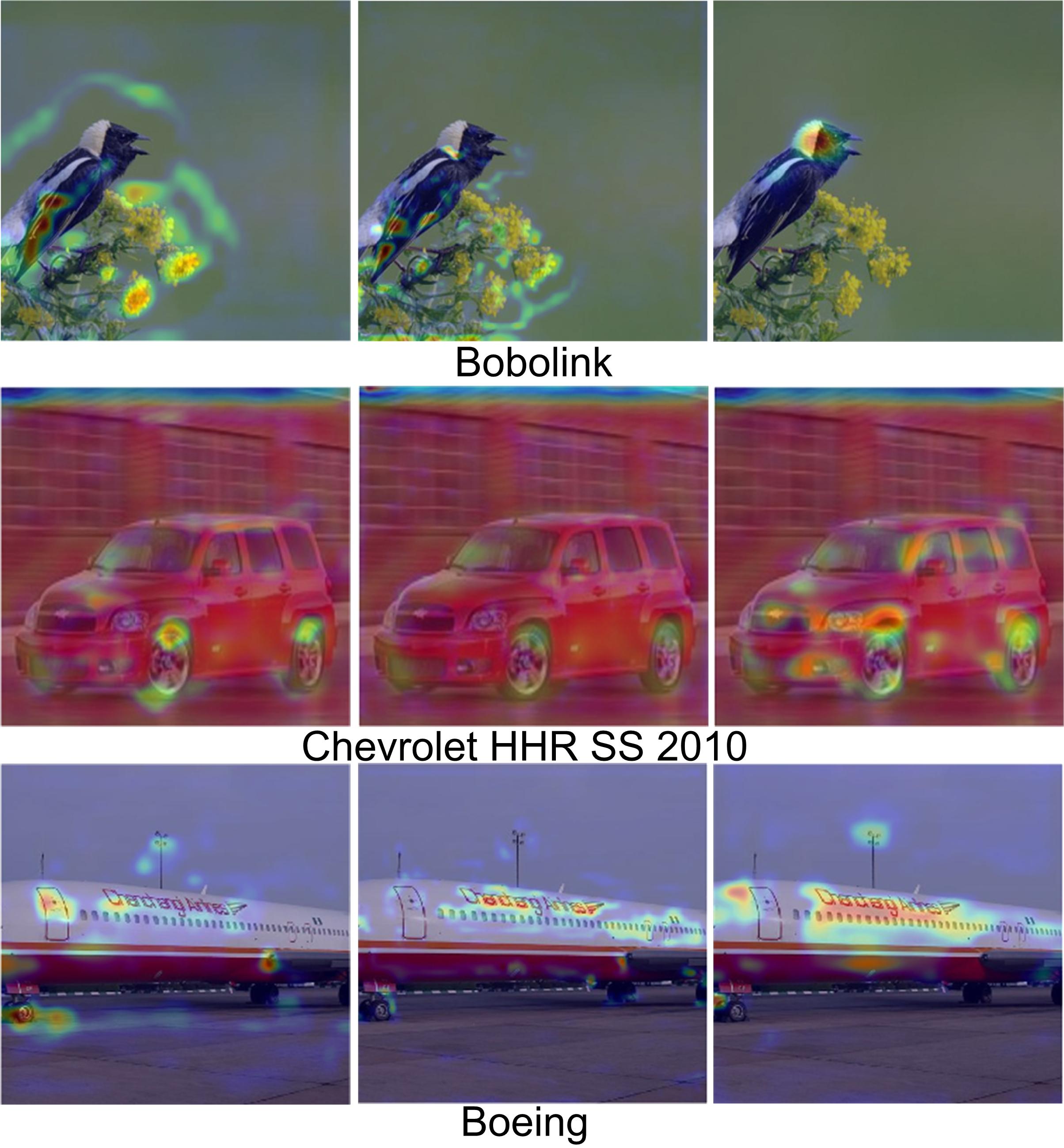}
\caption{Visualization of the responses in different branch routing modules.}
\vspace{-2mm}
\label{fig:routing}
\end{figure}

{\flushleft \textbf{Effectiveness of the attention transformer module.}} We construct a variant ``w/ Tree-Attn'', of the proposed ACNet model, to validate the effectiveness of the attention transformer module in Figure \ref{fig:ablation}. Specifically, we add the attention block in the transformer module in the ``w/ Tree'' method to construct the ``w/ Tree-Attn'' method. As shown in Figure \ref{fig:ablation}, the ``w/ Tree-Attn'' method performs consistently better than the ``w/ Tree'' method, producing higher top-1 accuracy with different number of channels, \ie, improving $0.4\%$ top-1 accuracy in average, which demonstrates that the attention mechanism is effective for FGVC. 

To further investigate the effect of ASPP module in our proposed model, we also conduct the ``w/ Tree-ASPP'' method, a variant of proposed ACNet model, where the only difference lies on between one convolution layer or ASPP module in the attention transformer module. As illustrated in Figure \ref{fig:ablation}, the attention transformer with ASPP module achieves better accuracy than the one with only one convolution layer. It indicates that the ASPP module improves the global performance by parallel dilated convolution layers with different dilated rates. Specifically, the ``w/ Tree-ASPP'' method improves $0.5\%$ top-1 accuracy in average. We can conclusion that multi-scale embedding and different dilated convolutions in ASPP module can facilitate helping the proposed tree network to obtain robust performance.

{\flushleft \textbf{Components in the branch routing module.}} We analyze the effectiveness of the global context block \cite{DBLP:journals/corr/abs-1904-11492} in the branch routing module in Figure \ref{fig:ablation}. Our ACNet method produces the best results with different number of channels in the branch routing module; while the top-1 accuracy drops $0.275\%$ in average after removing the global context block. Meanwhile, we also study the effectiveness of the pooling strategy in the branch routing module in Table \ref{tab:pooling}. We observe that using the global max-pooling (GMP) instead of the global average pooling (GAP) leads to $0.6\%$ top-1 accuracy drop on the CUB-200-2011 dataset. We speculate that the GAP operation encourages the filter to focus on high average response regions instead of the only maximal ones, which is able to integrate more context information for better performance.

{\flushleft \textbf{Coarse-to-fine hierarchical feature learning process.}}
The branch routing modules focus on different semantic regions (\eg, different object parts) or context information (\eg, background) at different levels, \eg, ${\rm R}^1_1$, ${\rm R}^2_1$, and ${\rm R}^2_2$ in Figure \ref{fig:structure}. As the example Bobolink shown in Figure \ref{fig:routing}, the ${\rm R}^1_1$ module focuses on the whole bird region at level-1; the ${\rm R}^2_1$ and ${\rm R}^2_2$ modules focus on the wing and head regions of the bird at level-2. As shown in the first row in Figure \ref{fig:leaf-node}, the four leaf nodes focus on several fine-grained object parts at level-3, \eg, different parts of the head region. In this way, our ACNet uses the coarse-to-fine hierarchical feature learning process to exploit discriminative features for more accurate results. This phenomenon demonstrates that our hierarchical feature extraction process in the tree $\mathbb{T}$ architecture gradually enforces our model to focus on more discriminative detail regions of object. 

\section{Conclusion}
In this paper, we present an attention convolutional binary neural tree (ACNet) for weakly supervised FGVC. Specifically, different root-to-leaf paths in the tree network focus on different discriminative regions using the attention transformer inserted into the convolutional operations along edges. The final decision is produced by max-voting the predictions from leaf nodes. The experiments on several challenging datasets show the effectiveness of ACNet. We present how we design the tree structure using coarse-to-fine hierarchical feature learning process in detail.

{\small
\bibliographystyle{ieee_fullname}
\bibliography{references}

\begin{thebibliography}{10}\itemsep=-1pt

\bibitem{DBLP:conf/wacv/AngelovaZL13}
Anelia Angelova, Shenghuo Zhu, and Yuanqing Lin.
\newblock Image segmentation for large-scale subcategory flower recognition.
\newblock In {\em WACV}, pages 39--45, 2013.

\bibitem{DBLP:journals/corr/BransonHBP14}
Steve Branson, Grant~Van Horn, Serge~J. Belongie, and Pietro Perona.
\newblock Bird species categorization using pose normalized deep convolutional
  nets.
\newblock {\em CoRR}, abs/1406.2952, 2014.

\bibitem{DBLP:conf/iccv/CaiZZ17}
Sijia Cai, Wangmeng Zuo, and Lei Zhang.
\newblock Higher-order integration of hierarchical convolutional activations
  for fine-grained visual categorization.
\newblock In {\em ICCV}, pages 511--520, 2017.

\bibitem{DBLP:journals/corr/abs-1904-11492}
Yue Cao, Jiarui Xu, Stephen Lin, Fangyun Wei, and Han Hu.
\newblock Gcnet: Non-local networks meet squeeze-excitation networks and
  beyond.
\newblock {\em CoRR}, abs/1904.11492, 2019.

\bibitem{DBLP:journals/pami/ChenPKMY18}
Liang{-}Chieh Chen, George Papandreou, Iasonas Kokkinos, Kevin Murphy, and
  Alan~L. Yuille.
\newblock Deeplab: Semantic image segmentation with deep convolutional nets,
  atrous convolution, and fully connected crfs.
\newblock {\em TPAMI}, 40(4):834--848, 2018.

\bibitem{DBLP:conf/ijcai/ChenLCWL18}
Tianshui Chen, Liang Lin, Riquan Chen, Yang Wu, and Xiaonan Luo.
\newblock Knowledge-embedded representation learning for fine-grained image
  recognition.
\newblock In {\em IJCAI}, pages 627--634, 2018.

\bibitem{Chen_2019_CVPR}
Yue Chen, Yalong Bai, Wei Zhang, and Tao Mei.
\newblock Destruction and construction learning for fine-grained image
  recognition.
\newblock In {\em CVPR}, pages 5157--5166, 2019.

\bibitem{DBLP:conf/cvpr/CuiZWLLB17}
Yin Cui, Feng Zhou, Jiang Wang, Xiao Liu, Yuanqing Lin, and Serge~J. Belongie.
\newblock Kernel pooling for convolutional neural networks.
\newblock In {\em CVPR}, pages 3049--3058, 2017.

\bibitem{DBLP:conf/eccv/DubeyGGRFN18}
Abhimanyu Dubey, Otkrist Gupta, Pei Guo, Ramesh Raskar, Ryan Farrell, and
  Nikhil Naik.
\newblock Pairwise confusion for fine-grained visual classification.
\newblock In {\em ECCV}, pages 71--88, 2018.

\bibitem{DBLP:conf/nips/DubeyGRN18}
Abhimanyu Dubey, Otkrist Gupta, Ramesh Raskar, and Nikhil Naik.
\newblock Maximum-entropy fine grained classification.
\newblock In {\em NeurIPS}, pages 635--645, 2018.

\bibitem{DBLP:journals/corr/abs-1711-09784}
Nicholas Frosst and Geoffrey~E. Hinton.
\newblock Distilling a neural network into a soft decision tree.
\newblock {\em CoRR}, abs/1711.09784, 2017.

\bibitem{DBLP:conf/cvpr/FuZM17}
Jianlong Fu, Heliang Zheng, and Tao Mei.
\newblock Look closer to see better: Recurrent attention convolutional neural
  network for fine-grained image recognition.
\newblock In {\em CVPR}, pages 4476--4484, 2017.

\bibitem{DBLP:journals/corr/GaoBZD15}
Yang Gao, Oscar Beijbom, Ning Zhang, and Trevor Darrell.
\newblock Compact bilinear pooling.
\newblock {\em CoRR}, abs/1511.06062, 2015.

\bibitem{DBLP:journals/jmlr/GlorotB10}
Xavier Glorot and Yoshua Bengio.
\newblock Understanding the difficulty of training deep feedforward neural
  networks.
\newblock In {\em AISTATS}, pages 249--256, 2010.

\bibitem{DBLP:conf/cvpr/HeZRS16}
Kaiming He, Xiangyu Zhang, Shaoqing Ren, and Jian Sun.
\newblock Deep residual learning for image recognition.
\newblock In {\em CVPR}, pages 770--778, 2016.

\bibitem{DBLP:conf/cvpr/HuSS18}
Jie Hu, Li Shen, and Gang Sun.
\newblock Squeeze-and-excitation networks.
\newblock In {\em CVPR}, pages 7132--7141, 2018.

\bibitem{DBLP:journals/corr/abs-1901-09891}
Tao Hu, Honggang Qi, Qingming Huang, and Yan Lu.
\newblock See better before looking closer: Weakly supervised data augmentation
  network for fine-grained visual classification.
\newblock {\em CoRR}, abs/1901.09891, 2019.

\bibitem{DBLP:conf/cvpr/HuangXTZ16}
Shaoli Huang, Zhe Xu, Dacheng Tao, and Ya Zhang.
\newblock Part-stacked {CNN} for fine-grained visual categorization.
\newblock In {\em CVPR}, pages 1173--1182, 2016.

\bibitem{DBLP:conf/icml/IoffeS15}
Sergey Ioffe and Christian Szegedy.
\newblock Batch normalization: Accelerating deep network training by reducing
  internal covariate shift.
\newblock In {\em ICML}, pages 448--456, 2015.

\bibitem{DBLP:conf/nips/JaderbergSZK15}
Max Jaderberg, Karen Simonyan, Andrew Zisserman, and Koray Kavukcuoglu.
\newblock Spatial transformer networks.
\newblock In {\em NeurIPS}, pages 2017--2025, 2015.

\bibitem{DBLP:journals/corr/abs-1804-02391}
Saumya Jetley, Nicholas~A. Lord, Namhoon Lee, and Philip H.~S. Torr.
\newblock Learn to pay attention.
\newblock {\em CoRR}, abs/1804.02391, 2018.

\bibitem{DBLP:journals/corr/JiaSDKLGGD14}
Yangqing Jia, Evan Shelhamer, Jeff Donahue, Sergey Karayev, Jonathan Long,
  Ross~B. Girshick, Sergio Guadarrama, and Trevor Darrell.
\newblock Caffe: Convolutional architecture for fast feature embedding.
\newblock {\em CoRR}, abs/1408.5093, 2014.

\bibitem{DBLP:journals/corr/KongF16}
Shu Kong and Charless~C. Fowlkes.
\newblock Low-rank bilinear pooling for fine-grained classification.
\newblock {\em CoRR}, abs/1611.05109, 2016.

\bibitem{DBLP:conf/cvpr/KrauseJYL15}
Jonathan Krause, Hailin Jin, Jianchao Yang, and Fei{-}Fei Li.
\newblock Fine-grained recognition without part annotations.
\newblock In {\em CVPR}, pages 5546--5555, 2015.

\bibitem{DBLP:conf/iccvw/Krause0DF13}
Jonathan Krause, Michael Stark, Jia Deng, and Li Fei{-}Fei.
\newblock 3d object representations for fine-grained categorization.
\newblock In {\em ICCVW}, pages 554--561, 2013.

\bibitem{DBLP:journals/corr/abs-1810-06058}
Haoming Lin, Yuyang Hu, Siping Chen, Jianhua Yao, and Ling Zhang.
\newblock Fine-grained classification of cervical cells using morphological and
  appearance based convolutional neural networks.
\newblock {\em CoRR}, abs/1810.06058, 2018.

\bibitem{DBLP:journals/corr/LinCY13}
Min Lin, Qiang Chen, and Shuicheng Yan.
\newblock Network in network.
\newblock In {\em ICLR}, 2014.

\bibitem{DBLP:conf/bmvc/LinM17}
Tsung{-}Yu Lin and Subhransu Maji.
\newblock Improved bilinear pooling with cnns.
\newblock In {\em BMVC}, 2017.

\bibitem{DBLP:conf/iccv/LinRM15}
Tsung{-}Yu Lin, Aruni {Roy Chowdhury}, and Subhransu Maji.
\newblock Bilinear {CNN} models for fine-grained visual recognition.
\newblock In {\em ICCV}, pages 1449--1457, 2015.

\bibitem{DBLP:journals/corr/LiuRB15}
Wei Liu, Andrew Rabinovich, and Alexander~C. Berg.
\newblock Parsenet: Looking wider to see better.
\newblock {\em CoRR}, abs/1506.04579, 2015.

\bibitem{DBLP:journals/corr/LiuXWL16}
Xiao Liu, Tian Xia, Jiang Wang, and Yuanqing Lin.
\newblock Fully convolutional attention localization networks: Efficient
  attention localization for fine-grained recognition.
\newblock {\em CoRR}, abs/1603.06765, 2016.

\bibitem{maji13fine-grained}
Subhransu Maji, Esa Rahtu, Juho Kannala, Matthew~B. Blaschko, and Andrea
  Vedaldi.
\newblock Fine-grained visual classification of aircraft.
\newblock {\em CoRR}, abs/1306.5151, 2013.

\bibitem{DBLP:conf/bmvc/MoghimiBSYVL16}
Mohammad Moghimi, Serge~J. Belongie, Mohammad~J. Saberian, Jian Yang, Nuno
  Vasconcelos, and Li{-}Jia Li.
\newblock Boosted convolutional neural networks.
\newblock In {\em BMVC}, 2016.

\bibitem{DBLP:journals/corr/PengHZ17}
Yuxin Peng, Xiangteng He, and Junjie Zhao.
\newblock Object-part attention driven discriminative localization for
  fine-grained image classification.
\newblock {\em CoRR}, abs/1704.01740, 2017.

\bibitem{DBLP:journals/ijcv/RussakovskyDSKS15}
Olga Russakovsky, Jia Deng, Hao Su, Jonathan Krause, Sanjeev Satheesh, Sean Ma,
  Zhiheng Huang, Andrej Karpathy, Aditya Khosla, Michael~S. Bernstein,
  Alexander~C. Berg, and Fei{-}Fei Li.
\newblock Imagenet large scale visual recognition challenge.
\newblock {\em IJCV}, 115(3):211--252, 2015.

\bibitem{DBLP:conf/iccv/SelvarajuCDVPB17}
Ramprasaath~R. Selvaraju, Michael Cogswell, Abhishek Das, Ramakrishna Vedantam,
  Devi Parikh, and Dhruv Batra.
\newblock Grad-cam: Visual explanations from deep networks via gradient-based
  localization.
\newblock In {\em ICCV}, pages 618--626, 2017.

\bibitem{DBLP:journals/corr/SimonyanZ14a}
Karen Simonyan and Andrew Zisserman.
\newblock Very deep convolutional networks for large-scale image recognition.
\newblock {\em CoRR}, abs/1409.1556, 2014.

\bibitem{DBLP:conf/eccv/SunYZD18}
Ming Sun, Yuchen Yuan, Feng Zhou, and Errui Ding.
\newblock Multi-attention multi-class constraint for fine-grained image
  recognition.
\newblock In {\em ECCV}, pages 834--850, 2018.

\bibitem{DBLP:conf/cvpr/SzegedyVISW16}
Christian Szegedy, Vincent Vanhoucke, Sergey Ioffe, Jonathon Shlens, and
  Zbigniew Wojna.
\newblock Rethinking the inception architecture for computer vision.
\newblock In {\em CVPR}, pages 2818--2826, 2016.

\bibitem{8805063}
Min Tan, Guijun Wang, Jian Zhou, Zhiyou Peng, and Meilian Zheng.
\newblock Fine-grained classification via hierarchical bilinear pooling with
  aggregated slack mask.
\newblock {\em Access}, 7:117944--117953, 2019.

\bibitem{DBLP:conf/icml/TannoAACN19}
Ryutaro Tanno, Kai Arulkumaran, Daniel~C. Alexander, Antonio Criminisi, and
  Aditya~V. Nori.
\newblock Adaptive neural trees.
\newblock In {\em ICML}, pages 6166--6175, 2019.

\bibitem{report-wahcub_200_2011}
C. Wah, S. Branson, P. Welinder, P. Perona, and S. Belongie.
\newblock {The Caltech-UCSD Birds-200-2011 Dataset}.
\newblock Technical report, California Institute of Technology, 2011.

\bibitem{DBLP:conf/iccv/WangSSZXZ15}
Dequan Wang, Zhiqiang Shen, Jie Shao, Wei Zhang, Xiangyang Xue, and Zheng
  Zhang.
\newblock Multiple granularity descriptors for fine-grained categorization.
\newblock In {\em ICCV}, pages 2399--2406, 2015.

\bibitem{DBLP:conf/cvpr/0004GGH18}
Xiaolong Wang, Ross~B. Girshick, Abhinav Gupta, and Kaiming He.
\newblock Non-local neural networks.
\newblock In {\em CVPR}, pages 7794--7803, 2018.

\bibitem{DBLP:journals/corr/WangCMD16}
Yaming Wang, Jonghyun Choi, Vlad~I. Morariu, and Larry~S. Davis.
\newblock Mining discriminative triplets of patches for fine-grained
  classification.
\newblock {\em CoRR}, abs/1605.01130, 2016.

\bibitem{DBLP:conf/cvpr/WangMD18}
Yaming Wang, Vlad~I. Morariu, and Larry~S. Davis.
\newblock Learning a discriminative filter bank within a {CNN} for fine-grained
  recognition.
\newblock In {\em CVPR}, pages 4148--4157, 2018.

\bibitem{DBLP:journals/corr/abs-1807-06521}
Sanghyun Woo, Jongchan Park, Joon{-}Young Lee, and In~So Kweon.
\newblock {CBAM:} convolutional block attention module.
\newblock {\em CoRR}, abs/1807.06521, 2018.

\bibitem{DBLP:journals/corr/abs-1712-05934}
Han Xiao.
\newblock {NDT:} neual decision tree towards fully functioned neural graph.
\newblock {\em CoRR}, abs/1712.05934, 2017.

\bibitem{DBLP:journals/corr/ZagoruykoK16a}
Sergey Zagoruyko and Nikos Komodakis.
\newblock Paying more attention to attention: Improving the performance of
  convolutional neural networks via attention transfer.
\newblock {\em CoRR}, abs/1612.03928, 2016.

\bibitem{DBLP:conf/cvpr/ZhangXEHZEM16}
Han Zhang, Tao Xu, Mohamed Elhoseiny, Xiaolei Huang, Shaoting Zhang, Ahmed~M.
  Elgammal, and Dimitris~N. Metaxas.
\newblock {SPDA-CNN:} unifying semantic part detection and abstraction for
  fine-grained recognition.
\newblock In {\em CVPR}, pages 1143--1152, 2016.

\bibitem{DBLP:conf/eccv/ZhangDGD14}
Ning Zhang, Jeff Donahue, Ross~B. Girshick, and Trevor Darrell.
\newblock Part-based r-cnns for fine-grained category detection.
\newblock In {\em ECCV}, pages 834--849, 2014.

\bibitem{DBLP:conf/iccv/ZhengFML17}
Heliang Zheng, Jianlong Fu, Tao Mei, and Jiebo Luo.
\newblock Learning multi-attention convolutional neural network for
  fine-grained image recognition.
\newblock In {\em ICCV}, pages 5219--5227, 2017.

\bibitem{DBLP:journals/corr/abs-1903-06150}
Heliang Zheng, Jianlong Fu, Zheng{-}Jun Zha, and Jiebo Luo.
\newblock Looking for the devil in the details: Learning trilinear attention
  sampling network for fine-grained image recognition.
\newblock {\em CoRR}, abs/1903.06150, 2019.

\end{thebibliography}
}

\end{document}